\documentclass[11pt,a4paper]{article}
\usepackage[hyperref]{emnlp-ijcnlp-2019}
\usepackage{times}
\usepackage{latexsym}
\usepackage{graphicx}
\usepackage{color}
\usepackage{kbordermatrix}
\usepackage{amsmath}
\usepackage{amssymb}
\usepackage{multirow}
\usepackage{url}
\usepackage{enumitem}

\aclfinalcopy

\title{NLNDE: Enhancing Neural Sequence Taggers with Attention and \\
Noisy Channel for Robust Pharmacological Entity Detection}

\author{Lukas Lange$^{1,2,3}$ \\
  \And
  Heike Adel$^1$ \\
  $^1$ Bosch Center for Artificial Intelligence, Renningen, Germany\\
  $^2$ Spoken Language Systems (LSV), Saarland University, Saarbr\"{u}cken, Germany\\
  $^3$ Saarbr\"{u}cken Graduate School of Computer Science, Saarbr\"{u}cken, Germany\\
  {\tt \{Lukas.Lange,Heike.Adel,Jannik.Stroetgen\}@de.bosch.com} \\
  \And
  Jannik Str\"{o}tgen$^1$ \\
  \\}

\date{}

\begin{document}
\maketitle

\begin{abstract}
Named entity recognition has been extensively studied on English news texts. However, the transfer to other domains and languages is still a challenging problem. In this paper, we describe the system with which we participated in the first subtrack of the PharmaCoNER competition of the BioNLP Open Shared Tasks 2019. Aiming at pharmacological entity detection in Spanish texts, the task provides a non-standard domain and language setting. However, we propose an architecture that requires neither language nor domain expertise. We treat the task as a sequence labeling task and experiment with attention-based embedding selection and the training on automatically annotated data to further improve our system's performance. Our system achieves promising results, especially by combining the different techniques, and reaches up to 88.6\% F1 in the competition. 
\end{abstract}

\section{Introduction}\label{sec:introduction}
The detection and classification of pharmacological and biomedical entities in texts is especially challenging due to the domain's nature with long and complex entity names, which usually requires the design and usage of handcrafted rules and features. Natural language processing (NLP) research focused on this topic for quite a while on English texts, e.g., the drugs and chemical names extraction challenge (CHEMDNER) \cite{st/CHEMDNER/Krallinger15} or tracks for chemical entity recognition at BioCreative \cite{st/BioCreative/Perez17}. Following these tasks, the Pharmacological Substances, Compounds and Proteins and Named Entity Recognition track (PharmaCoNER) is the first competition on this topic on Spanish data~\cite{pharmaconer2019}. 

\begin{figure}
	\centering
	\includegraphics[width=.48\textwidth]{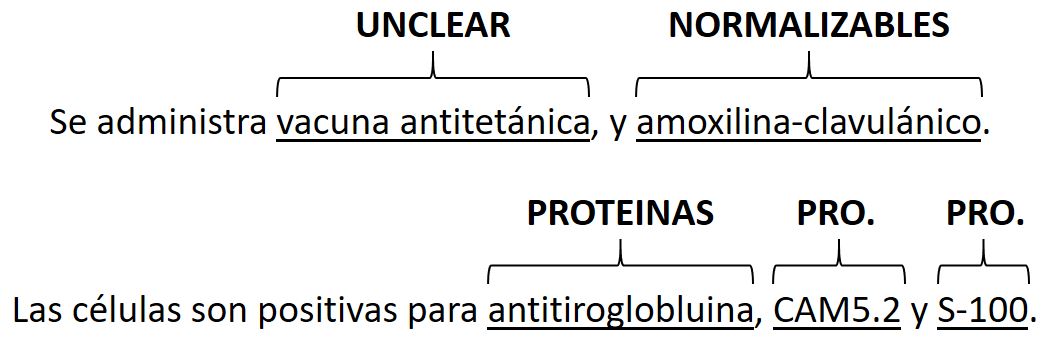}
	\caption{Annotated sample sentences (PRO. is short for \textsc{Proteinas}).}
	\label{fig:example}
\end{figure}

Named entity recognition (NER) and classification is the first subtrack of PharmaCoNER and aims at distinguishing four entity types: \textsc{Proteinas}, \textsc{Normalizables}, \textsc{No-Normalizables}, and \textsc{Unclear}. 
Our model was trained on all four entity types, although the \textsc{No-Normalizables} type was not considered during the official evaluation due to its ambiguous definition. Two annotated sample sentences from the training data are shown in Figure~\ref{fig:example}.

In this paper, we describe our submissions to and their results in the first subtrack of PharmaCoNER. We address this task as a sequence-labeling problem and implement a system that relies \textbf{N}either on \textbf{L}anguage \textbf{N}or on \textbf{D}omain \textbf{E}xpertise (NLNDE). For this, we use a combination of different state-of-the-art approaches from NLP to tackle its challenges without the need for handcrafted features. 

\begin{figure*}
	\centering
	\includegraphics[width=.96\textwidth]{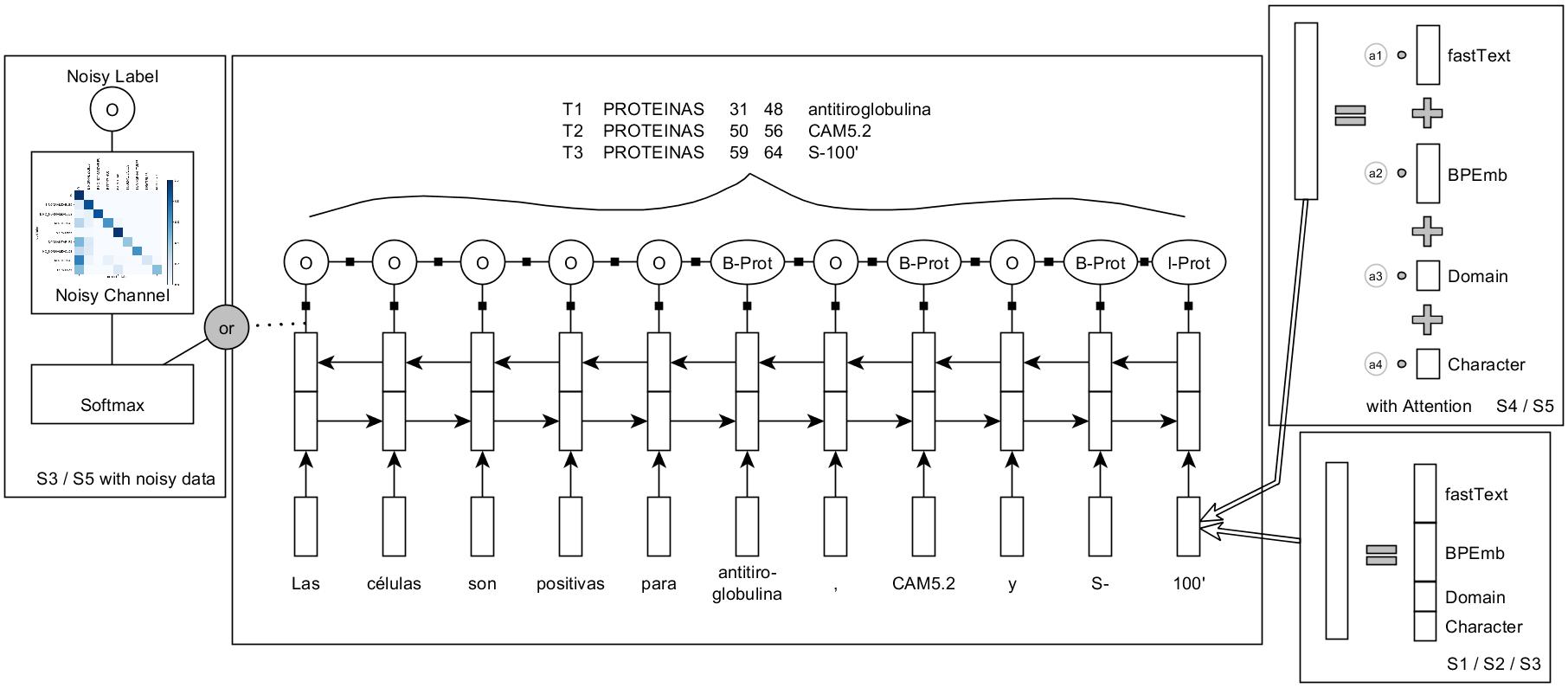}
	\caption{Architecture of our models. The label prefixes ``B-'' and ``I-'' show how we address the task as a sequence-labeling task. The word representations are either the concatenated embeddings (in runs S1--S3) or the attention-based weighted embeddings (in runs S4--S5).}
	\label{fig:model}
\end{figure*}

We train recurrent neural networks with conditional random field (CRF) output layers which are state of the art for different sequence labeling tasks, such as named entity recognition \cite{ner/Lample16}, part-of-speech tagging \cite{kemos2019} and de-identification \cite{dei/Liu17}. In our different runs, we further explore the advantages of domain-specific fastText embeddings that have been pre-trained on SciELO and Wikipedia articles \cite{emb/Soares19} to investigate the impact of domain knowledge. Note that the training of these embeddings requires only a collection of domain-specific text but no human domain expertise. Based on these models, we train an attention-based embedding selection function in order to leverage multiple different word embeddings effectively. Finally, we extend the training data with automatically annotated data, which was sampled from the same domain and annotated with information from Wikidata.\footnote{\url{https://www.wikidata.org/}}

\section{Methods}\label{sec:approach}
In this section, we present our system, the attention function for embedding selection, and the noisy channel model.  

\subsection{NLNDE System} \label{sec:system}
In Figure~\ref{fig:model}, the architecture of our models is depicted, which we explain in the following. 

\paragraph{Input Embeddings.}
We tokenize the input with the tokenizer provided by the shared task organizers~\cite{spaccc/Intxaurrondo19}. We noticed that the tokenizer sometimes merges multi-word expressions into a single token joined with underscores for contiguous words. As a result, some tokens cannot be aligned with the corresponding entity annotations. To address this, we split those tokens into their components in a postprocessing step. Then, we represent each token with the following embeddings (see bottom right box of Figure~\ref{fig:model}):
\begin{itemize}
	\item \emph{Character embeddings}: We use the concatenated last forward and backward hidden states of a bidirectional long short-term memory (BiLSTM) network \cite{lstm/Hochreiter97} over character embeddings (50 dimensions, randomly initialized, fine-tuned during training~\cite{ner/Lample16}).
	\item \emph{Domain-independent fastText embeddings} (100 dimensions, pre-trained on Spanish text~\cite{fastText/Grave18}). 
	\item \emph{Domain-specific fastText embeddings} (100 dimensions, pre-trained on Spanish SciELO and Wikipedia articles~\cite{emb/Soares19}).
	\item \emph{Byte-pair encoding embeddings} (300 dimensions, vocabulary size of 200,000, pre-trained on Spanish text~\cite{bpemb/heinzerling18}). 
\end{itemize}
Note that except for the character embeddings, we do not fine-tune any of the embeddings. All embeddings are concatenated into a single word representation vector.

\paragraph{Word Features.}\label{sub:features}
We also experiment with extending the input representations with the following features:

\begin{itemize}
	\item \emph{Part-of-speech (POS)}: The POS tags are generated by the POS-tagger provided by the shared task organizers~\cite{spaccc/Intxaurrondo19}. The tags are embedded into a 20-dimensional randomly initialized embedding and learned during training. The embedded vector is used as the representation for the POS tag.  
	\item \emph{Length}: For each word, we encode its length in a one-hot vector. Words with more than nine characters share the same vector (10 dimensions). 
	\item \emph{Frequency}: We consider the relative frequency $f$ of each word and bin the frequencies into ten groups. The first group contains the most frequent words that have relative frequencies above 1\% $(f>1\%)$.
	The remaining bins are constructed in the following manner: $f>0.5\%$, $f>0.1\%$, $f>0.05\%$, etc.
	(one-hot encoded, 10 dimensions). 
	\item \emph{Word shape}: We distinguish between uppercased, lowercased, starts with capital letter, numeric, mostly numeric, punctuation, mostly punctuation, only letters, alpha-numeric and other (one-hot encoded, 10 dimensions).
\end{itemize}
All features are concatenated into a single feature vector $f$ of 50 dimensions.

\paragraph{BiLSTM-CRF Layers.}
The input representation is fed into a BiLSTM with a conditional random field (CRF) output layer, similar to the model of \newcite{ner/Lample16}. The CRF output layer is a linear-chain CRF, i.e., it learns transition scores
between the output classes. For training, the forward algorithm is used to sum the scores for all possible sequences. During decoding, the Viterbi algorithm is applied to obtain the sequence with the maximum score.

\paragraph{Hyperparameters and Training.}
The hyperparameters are the same across all runs. We use a BiLSTM hidden size of 256 and train the network with the NADAM optimizer \cite{nadam/Dozat16} using a learning rate of 0.002 and a batch size of 32. For regularization, we employ early stopping on the development set and apply dropout with probability 0.5 on the input representations.

\subsection{Attention for Embedding Selection}\label{sub:attention}
As we are combining different word embeddings, some of them may be more beneficial for certain words than others, 
e.g., domain-specific embeddings for in-domain words. \newcite{emb/dynamic/Kiela18} used an attention mechanism for weighting and selecting the best embeddings for each word. We extend this idea and propose the following attention function to weight the embeddings depending on additional word features.

For the attention-based models, all $n$ embeddings $e$ are mapped to the same size using a linear mapping $Q_i \in \mathbb{R}^{E \times E_i}$ without bias, with $x_i \in \mathbb{R}^E$ being the $i$-th embedding $e_i$ mapped from their original size $E_i$ to the maximal embedding size $E = \max_m (E_m)$. 
\begin{align}
x_i = Q_i \cdot e_i
\end{align}

In order to allow the model to make an informed decision which embeddings to focus on, we use the word features described in Section \ref{sub:features} as an additional input to the attention function. The vector $f \in \mathbb{R}^F$ representing the features for each word is concatenated to each embedding $x_i$. The attention weight $a_i$ for each embedding $x_i$ is computed with the softmax function, by feeding $x_i$ and $f$ into a fully-connected hidden layer of size $H$ with the parameters 
$W \in \mathbb{R}^{H \times E}$,
$U \in \mathbb{R}^{H \times F}$,
$V \in \mathbb{R}^{1 \times H}$.
\begin{align}
a_i = \frac{\exp(V \cdot \tanh(W x_i + U f))}{\sum_{l=1}^n \exp(V \cdot \tanh(W x_l + U f))}
\end{align}

Finally, the embeddings $x_i$ are weighted using the attention weights $a_i$ resulting in the word representation: 
\begin{align}
e = \sum_i a_i \cdot x_i
\end{align}

Then, this word representation $e \in \mathbb{R}^E$ is fed into the BiLSTM-CRF. Compared to a concatenation of the different embeddings, this results in a lower-dimensional word representation and, thus, requires fewer parameters in the BiLSTM layer. The attention-based embedding selection is shown in the upper right box of Figure~\ref{fig:model}. 

\subsection{Training on Noisy Data}
\label{sub:noise}
As it was shown in multiple low-resource settings~\cite{noise/Dgani2018,noise/Fang16,noise/Mnih2012,noise/Paul19,noise/Yang18}, the performance of NER and other NLP systems can be substantially improved by training on additional noisy data which is labeled in a distantly supervised manner~\cite{distant/Mintz09}. With this approach, the noisy data is cheap to create, but also error-prone and can even decrease performance if used as training data without noise handling as shown by \newcite{noise/Hedderich18}.

\paragraph{Extraction of Noisy Data.}
We create gazetteers for the different entity types by extracting names and aliases of possible entities from Wikidata for the following categories and their subclasses:\footnote{WikiData identifiers used for the extraction: Q8047, Q7187, Q11364, Q8054, Q81915, Q37756, Q8066, Q79460, Q11358, Q177719, Q189720, Q11367, Q7946, Q28745, Q42962,	Q2356542, Q47154513, Q172847, Q756, Q81163, Q134808.}

\begin{itemize}
\item
\textsc{Proteinas}:
	enzyme,
	gene,
	hormone,
	protein.
\item
\textsc{Normalizables}:
	allotropy,
	alloy,
	amino acid,
	antibody,
	carbohydrate,
	diagnostic procedure,
	dye,
	lipid,
	mineral,
	nucleotide,
	oil,
	reagent,
	chemical compounds,
	peptide,
	plant,
	polymer,
	vaccine. 
\item
\textsc{Unclear} and \textsc{No-Normalizables}:
The gazetteer was constructed from entity mentions in the training data that appeared at least twice and examples from the annotation guidelines. 
\end{itemize}
Then, we retrieve unlabeled documents from the same domain from the SciELO archive~\cite{scielo/packer98}. Finally, we use the extracted gazetteers to automatically annotate the SciELO data with the method from \citet{noise/Lange19}.
We use case-insensitive string matching for \textsc{Proteinas} and strict string matching for the other types. This allows to create additional training instances, but at the same time introduces noise into the system. To avoid that the noisy labels result in a decrease of performance, we train on the noisy data with a special noise handling method adapted from \newcite{noise/Goldberger16}, which will be described in the following. 

\begin{figure}
	\centering
	\includegraphics[width=.4\textwidth]{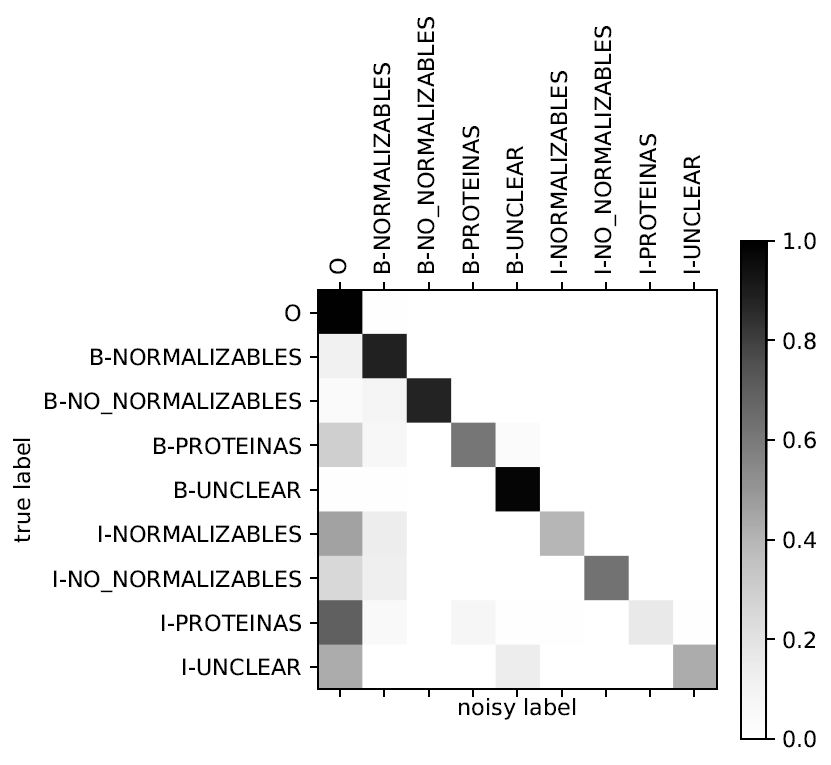}
	\caption{Confusion matrix for the automatic annotation on the training data used for the noisy channel initialization.}
	\label{fig:matrix}
\end{figure}

\paragraph{Noisy Channel and Confusion Matrix.}
First, we annotate each word of the training data using the same method as for generating the noisy data. Thus, each word in the training data has a clean, true label $y$ and a noisy label $\hat{y}$ from which we can model the noise distribution $p(\hat{y}=j|y=i)$ with a confusion matrix, as shown in Figure~\ref{fig:matrix}. We transform the distribution of the predicted (clean) labels to the noisy label distribution through a so-called noisy channel \cite{noise/Goldberger16}:
\begin{align}
p(\hat{y}=j|x) = \sum^k_{i=1} p(\hat{y}=j| y=i)p(y=i|x) \label{eq:noise-prediction}
\end{align}
where $k$ is the number of classes and $p(y=i|x)$ is the probability of a label $y$ having a specific class $i$ given the feature $x$.

We initialize the noisy channel weights using the learned confusion matrix on the training set, for which clean and noisy labels are available. 

\paragraph{Training with Confusion Matrix.}
The sequence tagging model is then trained alternately on the clean data with the CRF output layer and on the noisy data with the noisy channel layer, as shown in Figure~\ref{fig:model}. The number of noisy training instances is constantly decreased by 5\% after every training epoch to at least 100 sentences, as we observed that the noisy data helps in particular for the first epochs, but decreases performance if the amount is not reduced. Note that we shuffle the noisy data after each training epoch. Thus, the model is trained on new samples of noisy sentences in every epoch.

\section{Submissions}\label{sec:submissions}
We submitted five runs to the PharmaCoNER competition. All of them are based on the architecture described in Section~\ref{sec:system}. 

\begin{itemize}
	\item[S1]
	(\textit{Base}):  Our first run, the base system for all of the following runs, uses a concatenation of three  embeddings (character, BPEmb, fastText) which were all trained on Wikipedia. Thus, this run does not include any form of domain knowledge, and it uses neither noisy data nor attention for embedding selection. 
	\item[S2] 
	(\textit{Domain}): 
	Our second run uses the three embeddings from S1 plus domain-specific fastText embeddings to incorporate knowledge about word distributions within the domain. 
	\item[S3] 
	(\textit{Noise}):
	Our third run extends the model of S2 with training on additional noisy data (cf.\  Section \ref{sub:noise}).
	Moreover, we use the feature vector as an additional input, which is different from runs S1 and S2. 
	\item[S4] 
	(\textit{Attention}):
	The fourth run uses the attention function for word embedding selection (cf. Section~\ref{sub:attention}).
	Apart from that, the model is identical to S2 and only trained on clean data.
	\item[S5] 
	(\textit{Attention+Noise}): 
	Our last run has the same architecture as S4 but is trained on the noisy data in addition. 
	It thus combines
	domain-independent and domain-specific word embeddings,
	attention-based embedding selection, and training on noisy data. 
\end{itemize}
\section{Results and Analysis}\label{sec:results}
This section describes our results and analysis. 

\subsection{Experimental Results}
\begin{table}
	\centering
	\footnotesize
	\begin{tabular}{l|ccc|ccc} 
		& 
		\multicolumn{3}{|c|}{\textit{Development}} &
		\multicolumn{3}{|c}{\textit{Test}} \\ 
		{\textbf{S$_{ID}$}} & P & R & F1 & P & R & F1  \\ \hline
		S1 & 86.3 & 85.0 & 85.6 & 85.5 & 85.3 & 85.4 \\
		S2 & 87.1 & 85.8 & 86.4 & 86.3 & 85.9 & 86.1 \\
		S3 & 87.5 & 87.5 & 87.5 & 85.1 & 86.0 & 85.6 \\
		S4 & 88.0 & \bf 88.0 & 88.0 & 85.2 & 87.2 & 86.2 \\
		S5 & \bf 89.1 & 87.5 & \bf 88.3 & \bf 89.0 & \bf 88.3 & \bf 88.6
	\end{tabular}
	\caption{Precision (P), Recall (R) and F1 for Task 1.}
	\label{tab:results}
\end{table}

In Table~\ref{tab:results}, we report the results on the PharmaCoNER development and test sets using the official shared task evaluation metrics. 

Adding domain-knowledge (S2) to the base model S1 improves the performance on the development and the test set. The training on noisy data (S3) and the attention function alone (S4) do not lead to strong improvements on the test set; the noise model S3 even decreases performance. 
The combination of all proposed methods (run S5 \textit{Attention+Noise}) outperforms all other models. 

While we are able to see the improvements step by step introduced by our methods on the development set, such improvements are not observable one-to-one on the test set. We assume that model S5 performs best at generalizing to unseen words due to the training on additional data and the attention function based on basic word properties like word length or frequency. 
The other models seem to overfit on the development set, even though this set was never used for training but only for early stopping. 

\subsection{Analysis of Attention Weights}
\begin{figure}
	\centering
	\vspace{-3mm}
	\includegraphics[width=.4\textwidth]{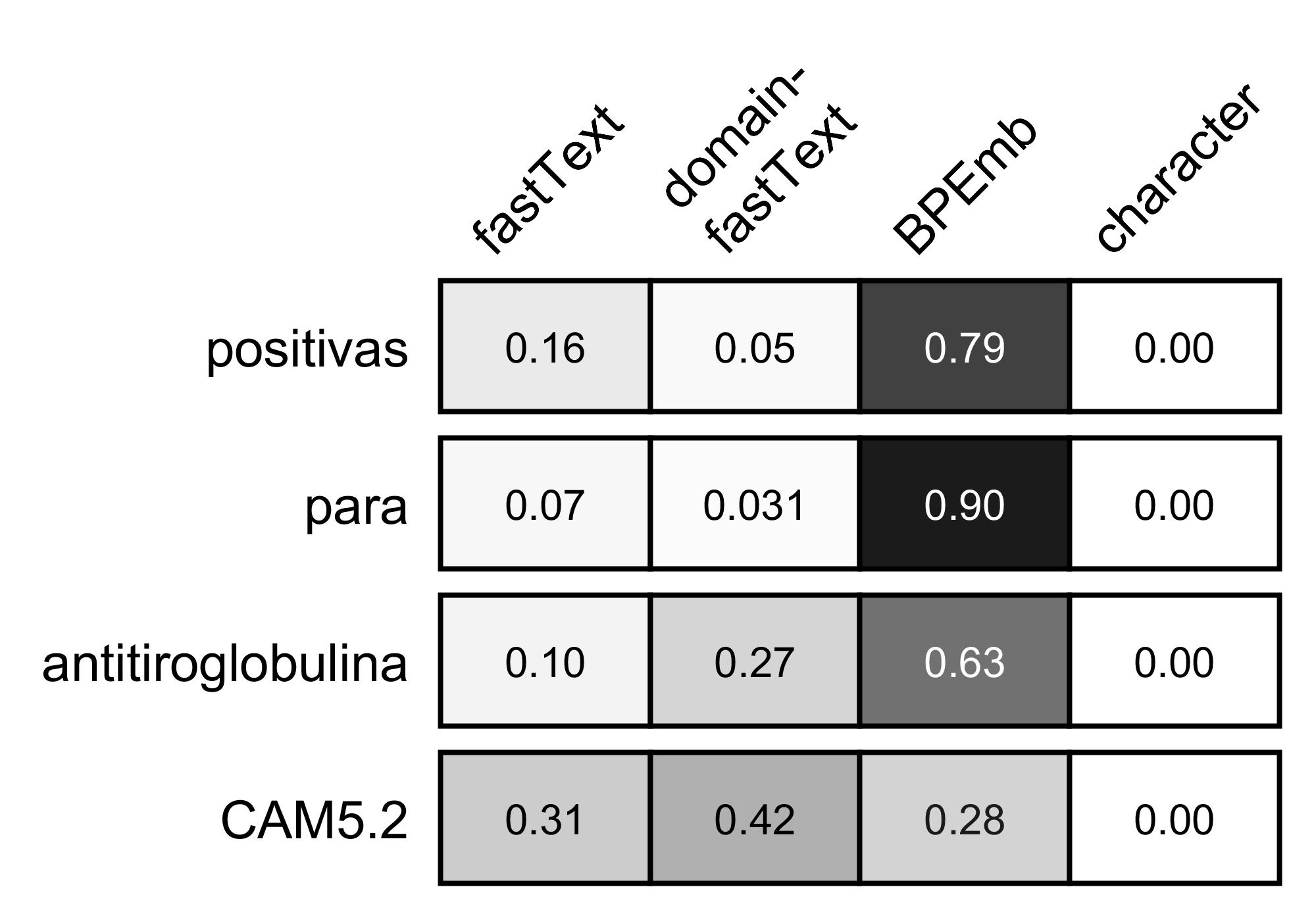}
	\caption{The attention weights of our model for the four embeddings. Darker color indicates higher weight.}
	\label{fig:attention}
\end{figure}

The attention-based models learn to focus mostly on the byte-pair-encoding embeddings, as shown in Figure~\ref{fig:attention}.  In particular, for words from the general domain (positivas) and stopwords (para), our model focuses on these embeddings. For domain-specific words (antitiroglobulina, CAM5.2), the model learns to focus more on the fastText embeddings and especially the domain-specific embeddings. Interestingly, the character embeddings are never assigned a noticeable weight. This may be attributed to the fact that the other embeddings are all subword embeddings and that they are able to generate meaningful vectors for out-of-vocabulary words. Moreover, the character embeddings were randomly initialized and had to be learned during training while the other models were pretrained. 

\section{Conclusions}\label{sec:conclusion}
In this paper, we described our system for the first subtrack of the PharmaCoNER competition. We trained a bi-directional long short-term memory network and explored different input representations. We proposed to use a feature-based attention function for embedding selection and training on noisy data, which in combination increased performance by more than 3 F1 points up to 88.6\%. This shows that we can successfully extract these special types of entities without the need for domain or language-specific model architectures.

\section*{Acknowledgments}
We would like to thank the anonymous reviewers for their helpful suggestions and comments.

\bibliography{2019-BioNLP}
\bibliographystyle{acl_natbib}

\end{document}